\documentclass{article}
\usepackage[letterpaper,textwidth=5.5in,textheight=9in,centering]{geometry}
\usepackage[T1]{fontenc}
\usepackage{times}
\usepackage{needspace}
\usepackage[authoryear,round]{natbib}
\setcitestyle{citesep={;},aysep={,},yysep={;}}
\date{}
\usepackage{amsmath,amsfonts,bm}

\def\eqref#1{equation~\ref{#1}}
\def\1{\bm{1}}

\DeclareMathAlphabet{\mathsfit}{\encodingdefault}{\sfdefault}{m}{sl}
\SetMathAlphabet{\mathsfit}{bold}{\encodingdefault}{\sfdefault}{bx}{n}

\usepackage{amsmath,amsfonts,amssymb,amsthm,bm}
\usepackage{graphicx}
\usepackage{algorithm}
\usepackage{algpseudocode}
\usepackage{tikz}
\usetikzlibrary{arrows.meta,positioning,fit,calc}
\usepackage{multirow}
\usepackage{array}
\usepackage{booktabs}
\usepackage{hyperref}
\usepackage{url}
\hypersetup{colorlinks=false,pdfborder={0 0 0}}

\newcommand{\FIGURE}[3]{\centering #1\caption{#2}#3}
\newcommand{\TABLE}[3]{\caption{#1}\centering
  \resizebox{\textwidth}{!}{#2}
  \par\smallskip{\footnotesize #3}}
\newcommand{\up}{}

\title{Sequence-Structured Decision Learning for Branch-and-Bound in Mixed-Integer Linear Programming}

\author{\parbox{\dimexpr\textwidth-2\tabcolsep\relax}{\centering
Tengfei Liu$^{1}$ \quad Jinrun Liu$^{1}$ \quad Bai Yan$^{2}$\\[0.4ex]
Yuhui Shi$^{1}$ \quad Qi Zhao$^{1}$\thanks{Corresponding author: Qi Zhao (\texttt{zhaoq@sustech.edu.cn}).}\\[1ex]
\small $^{1}$Department of Computer Science and Engineering\\
\small Southern University of Science and Technology, Shenzhen, China\\[0.4ex]
\small $^{2}$School of Computer Science and Technology\\
\small Dongguan University of Technology, Dongguan, China\\[0.7ex]
\small\texttt{\{12332470,12432683\}@mail.sustech.edu.cn}\\
\small\texttt{yanbai@dgut.edu.cn} \quad
\small\texttt{\{shiyh,zhaoq\}@sustech.edu.cn}
}}

\begin{document}

\maketitle

\begin{abstract}
Branch-and-bound (BnB) solves mixed-integer linear programs by constructing and exploring a search tree. Node selection and branching are two recurrent decisions that directly determine the tree. Learning-based approaches typically optimize either decision in isolation and condition on a mostly local view, leaving the coupled evolution of the search tree insufficiently modeled. This paper proposes Seq-BnB, a sequence-structured framework that learns branching and node selection from a shared global representation of the evolving BnB process. Seq-BnB serializes node creations, node updates, node-selection events, and branching events into typed tokens, producing a causal event sequence of the search trajectory. A self-attention encoder summarizes this global search history, a bipartite graph neural network encodes the current LP relaxation, and context-conditioned decision heads score branching variables and open nodes. To reduce distribution shift between expert demonstrations and learned policies, Seq-BnB is trained with DAgger-style data aggregation. On four MILP benchmark families, Seq-BnB consistently processes fewer BnB nodes than heuristic and learning-based baselines, with strong solution times on both standard and larger transfer instances. While this paper evaluates Seq-BnB instantiation for branching and node selection, it accommodates additional solver decisions, such as cut selection, by introducing new action tokens and decision heads.
\end{abstract}


\section{Introduction}\label{sec:introduction}
Mixed-integer linear programming (MILP) is a central modeling tool in operations research and decision analytics, with applications ranging from network design and scheduling to allocation and logistics \citep{nemhauser1988integer,wolsey1998integer}. The classical Branch-and-Bound (BnB) algorithm solves MILPs by recursively decomposing the feasible region into a dynamically growing search tree \citep{land1960automatic,dakin1965tree}. Modern branch-and-cut solvers enrich this framework with presolving, cutting planes, primal heuristics, and sophisticated tree management \citep{achterberg2009scip,gamrath2020scip}. Within the process, two recurrent decisions directly govern the tree growth: node selection chooses the open subproblem to process next, and branching partitions that subproblem into new descendants \citep{linderoth1999computational,achterberg2005branching}. Their interaction influences when incumbents are discovered, how bounds improve, which subtrees are pruned, and ultimately how much search is required.

A large body of work has studied these decision rules. Classical branching rules include strong branching, pseudocost branching, and reliability branching; search rules include best-estimate search and related node-selection strategies \citep{achterberg2005branching,achterberg2009scip,lodi2017learning}. Learning-based methods replace these hand-designed scores with policies learned from solver data. For branching, they imitate strong branching or incorporate search-tree information to rank candidate variables at the current node \citep{khalil2016learning,alvarez2017machine,gasse2019exact,zarpellon2021parameterizing,gupta2022lookback,bengio2021machine}. 
For node selection, they rank or compare open nodes in the active set \citep{he2014learning,labassi2022learning}. 
Reinforcement learning and retrospective trajectory construction were also explored for branching and BnB search control \citep{etheve2020reinforcement,parsonson2023reinforcement}. Most methods learn one decision type at a time: branching models choose variables, whereas node-selection models choose open nodes.

Recent work moves beyond isolated solver decisions. Multi-task and foundation-model approaches learn shared MILP representations for different prediction or control tasks, while unified BnB policies jointly consider node selection and variable selection \citep{cai2024multi,li2024towards,du2025learning}. Other recent methods incorporate branching history or sequential structure, or explicitly represent the BnB tree for node selection \citep{lin2022tbrant,zhang2026mamba,zhang2025trgnn}. These studies demonstrate the value of shared representations and coupled solver decisions. They also leave open how to represent the chronological history of node-selection and branching events as a common context for both decisions.

The remaining gap is therefore not only whether multiple solver decisions are learned together, but how the evolving BnB process is represented. Branching and node selection act on different candidate sets but modify the same search tree. A branching policy observes fractional variables at the current LP relaxation, while a node-selection policy observes open subproblems in the active tree. Even when the two decisions are jointly modeled, an effective policy must account for the trajectory that has produced the current tree, including previous branching choices, search-order decisions, incumbent updates, bound changes, and the evolving distribution of open nodes. Without such a unified historical representation, a locally favorable decision for one component may lead to an unfavorable downstream tree when evaluated together with the other component.

The sequential nature of BnB further strengthens this issue. BnB is a policy-induced process: deployed decisions change the nodes, candidate variables, and search histories encountered later. This distributional issue is well known in imitation learning and sequential prediction \citep{ross2011reduction}, and it is particularly important inside BnB because early actions reshape the subsequent search states. Joint decision learning therefore requires two ingredients: a shared representation of the evolving search process and training data collected from states induced by the learned policy.

This motivates us to propose Seq-BnB in this paper. Seq-BnB provides a shared, causally available search-state representation for coupled BnB decisions. Rather than treating the BnB tree only as a collection of locally scored candidates, Seq-BnB incrementally serializes the evolving tree and its decision trajectory into typed tokens. Node tokens describe created or updated subproblems, while action tokens record node-selection and branching events together with global search statistics. A self-attention encoder \citep{vaswani2017attention} summarizes the resulting global context. For the current LP relaxation, a bipartite graph neural network encodes local variable--constraint structure, following the representation used in learning-based branching \citep{gasse2019exact}. Cross-attention combines global tree context with local candidate features and produces both branching and node-selection scores. To mitigate distribution shift caused by deploying learned decisions, Seq-BnB is trained with a DAgger-style data aggregation procedure that progressively incorporates states visited by the learned policy.

The sequential formulation is not tied to only branching and node selection decisions. These two form the minimal coupled decision set studied in this paper because they determine where and how the BnB tree grows. More broadly, learning can also be applied to other solver components, including primal solution prediction, cut selection, primal heuristics, heuristic scheduling, and solver configuration \citep{nair2020solving,sonnerat2021learning,paulus2022learning,deza2023machine,chmiela2021learning}. The typed-token representation offers a common interface for such actions: additional solver events can be recorded as new token types and equipped with corresponding prediction heads. This perspective is also complementary to emerging generalist and foundation-model approaches for MILP \citep{li2024towards}. We regard this extensibility as a design implication of the representation, while empirically validating the branching-and-node-selection instantiation in this paper.

To summarize, our contributions are as follows:
\begin{itemize}
    \item We formulate branching and node selection as coupled tree-construction decisions in BnB. Instead of treating branching as a local variable-ranking problem and node selection as a separate open-node-ranking problem, we model them as two recurrent actions that jointly induce the search tree. 
    \item We introduce a causal event-sequence representation for BnB. Seq-BnB serializes node creations, node-state updates, node-selection actions, and branching actions into typed tokens. The resulting sequence provides a common global state for heterogeneous solver decisions while preserving the online execution order of BnB. This representation differs from local LP-graph encodings by explicitly modeling search history, tree evolution, and previous solver actions.
    \item We develop a joint learning architecture for branching and node selection. A shared self-attention encoder summarizes the global event sequence, a bipartite graph neural network represents the current LP relaxation, and context-conditioned decision heads score both branching variables and open nodes. The two heads are trained with a joint objective and DAgger-style data aggregation, allowing the policy to learn from states induced by its own deployment inside the solver.
\end{itemize}

We provide computational studies on four MILP benchmark families. Seq-BnB consistently processes fewer BnB nodes than heuristic and learning-based baselines on both standard and larger transfer instances. Ablation studies show that removing sequence context from the branching head, learning only one of the two decisions, or training without DAgger substantially degrades performance. These results support the central claim that shared sequential context improves coupled BnB decision making.

\section{Background and Related Work}\label{sec:related_work}
A MILP has continuous and integer variables coupled by linear constraints. BnB solves LP relaxations over subproblems and prunes nodes that are infeasible, integer feasible, or unable to improve the incumbent. At step $t$, the solver selects an open node $n_t\in\mathcal O_t$; if branching is needed, it selects a fractional integer variable $x_{j_t}$ and creates child subproblems. Because branching changes future open nodes and node selection determines where branching is applied, the two decisions jointly induce the evolving tree. Classical policies include strong branching, pseudocost branching, and best-estimate search \citep{achterberg2005branching,linderoth1999computational}.

Learning-based MILP research has mainly targeted individual BnB decisions \citep{lodi2017learning,bengio2021machine}. Branching policies progressed from hand-crafted features and strong-branching imitation to bipartite graph representations and reinforcement learning \citep{khalil2016learning,alvarez2017machine,gasse2019exact,parsonson2023reinforcement}. Recent work also models branching history: T-BranT uses a tree-aware transformer \citep{lin2022tbrant}, while Mamba-Branching models sequences of branching states and decisions \citep{zhang2026mamba}. Node-selection policies learn search orders or compare open nodes \citep{he2014learning,labassi2022learning}; TRGNN represents the evolving tree as a tripartite graph for node selection \citep{zhang2025trgnn}. These history- and tree-aware methods learn one decision type at a time.

Shared MILP representations have also been investigated across solver tasks and problem classes \citep{cai2024multi,li2024towards}. More closely related, Du et al.~\citep{du2025learning} jointly integrate node and variable selection, recognizing their interdependence. Seq-BnB focuses on a shared chronological sequence of node creations, updates, node-selection actions, and branching actions as context for both decisions. This representation combines the evolving search history with the current LP graph and supports on-policy data aggregation. Further background and comparisons appear in Appendix~\ref{app:related_work}.

\section{Seq-BnB Framework}\label{sec:method}
\begin{figure}
\FIGURE
{\includegraphics[width=1.0\textwidth]{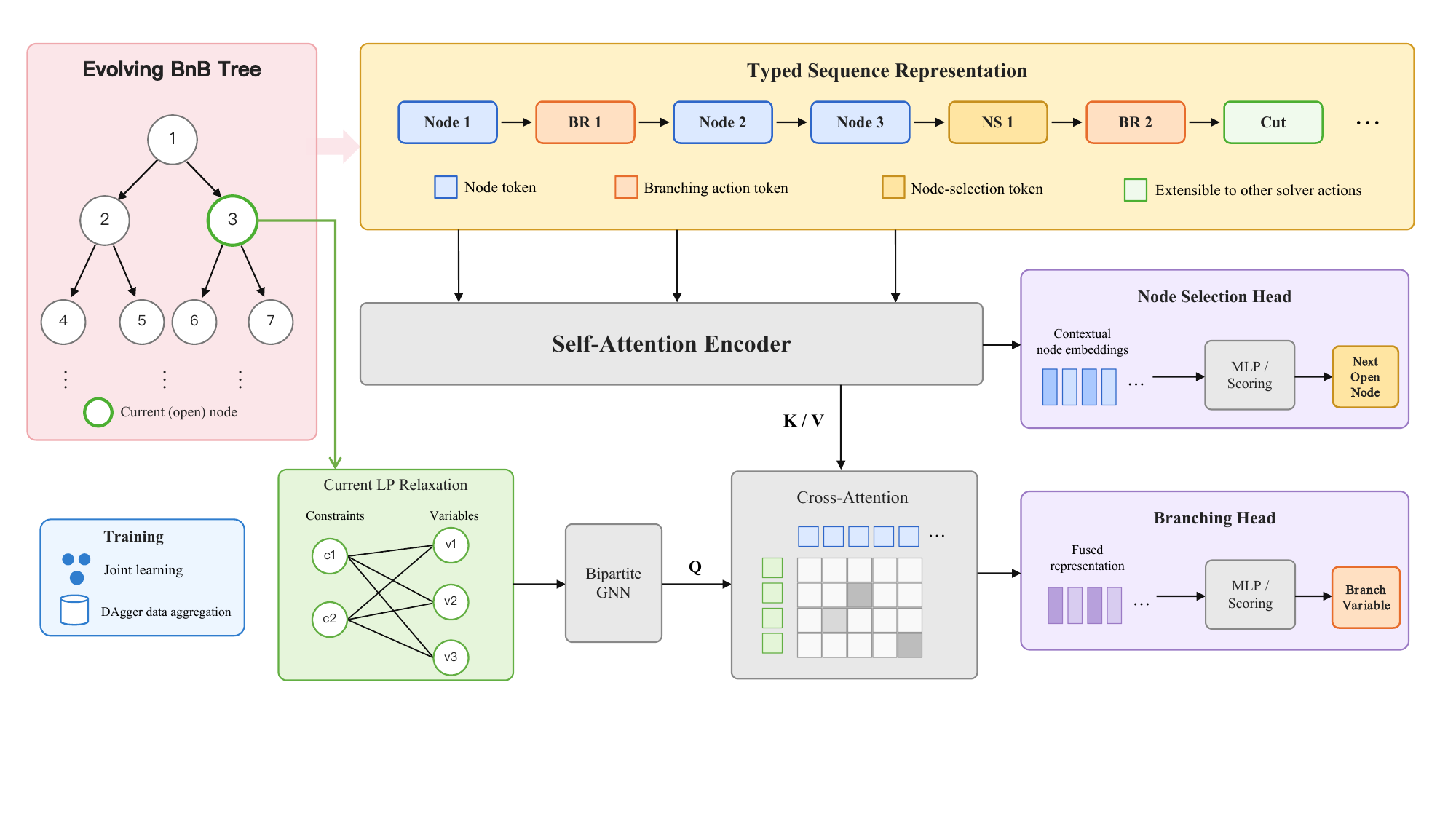}}
{Architecture of Seq-BnB. The evolving BnB tree and solver actions are serialized into typed tokens and encoded by a self-attention module. Local LP relaxations are encoded by a bipartite GNN, and cross-attention conditions branching and node-selection decisions on the global search context. \label{fig:framework}}
{}
\end{figure}

Seq-BnB has three components: a sequence-structured global representation of the evolving BnB tree, a local graph encoder for the current LP relaxation, and decision heads that query the shared encoded context. In the present instantiation, the heads correspond to branching and node selection, the two coupled decisions that generate the search tree. Figure \ref{fig:framework} summarizes the architecture.

\subsection{Sequential Representation of BnB}\label{sec:seq_rep}
Let $\mathcal{Z}_t=(z_1,z_2,\ldots,z_{L_t})$ denote the sequence available immediately before decision step $t$. Each token belongs to one of three types:
\begin{equation}
 z_i\in \{\textsc{Node},\textsc{NS},\textsc{BR}\},
\end{equation}
where \textsc{Node} records the creation or state update of a BnB node, \textsc{NS} records a node-selection decision, and \textsc{BR} records a branching decision. This event-based representation is aligned with the solver's natural execution order and can be updated online as the search evolves. Conceptually, the sequence is an incremental serialization of the generated search tree together with the actions that produced its growth, providing a global carrier of information shared across solver decisions. In implementation, the bounded window described below retains recent history and all currently open nodes, avoiding the claim that every historical token must be retained indefinitely.

The token vocabulary is modular. For example, a future extension that learns cut selection could introduce a \textsc{Cut} action token recording the cut-selection state and selected cut, and attach a cut-scoring head to the same encoded context. Thus, the representation is designed as an interface for decision learning over the BnB process; the current paper focuses on the \textsc{NS} and \textsc{BR} tokens and heads.

For a BnB node $u$, the node-token feature vector is
\begin{equation}
\mathbf{x}^{\textsc{node}}(u)=
[\ell(u),c(u),d(u),\mathbf{1}_{\mathrm{visited}}(u),\mathbf{1}_{\mathrm{fathomed}}(u),
\mathbf{1}_{\mathrm{incumbent}}(u),s(u),g_t],
\label{eq:node_token}
\end{equation}
where $\ell(u)$ is the LP lower bound, $c(u)$ is the estimate value, $d(u)$ is depth, $s(u)$ is the creation or visit order, and $g_t$ is a global gap indicator. The binary indicators record whether the node has been processed, fathomed, or associated with an incumbent update.

After the node-selection decision has selected $n_t$ from $\mathcal O_t$ and the LP relaxation at $n_t$ has been solved, we record the node-selection event by constructing
\begin{equation}
\mathbf{x}^{\textsc{ns}}_t=
[g_t,|\mathcal{O}_t|,t,d(n_t),\ell(n_t),c(n_t),\mathbf{s}^{\mathrm{node}}(n_t)],
\label{eq:ns_token}
\end{equation}
where $\mathbf{s}^{\mathrm{node}}(n_t)$ is a pooled summary of the local graph representation at $n_t$. 

For branching, if variable $x_{j_t}$ is selected at $n_t$, we construct
\begin{equation}
\mathbf{x}^{\textsc{br}}_t=
[g_t,|\mathcal{O}_t|,t,d(n_t),\ell(n_t),c(n_t),\mathbf{h}^{\mathrm{var}}_{j_t}(n_t)],
\label{eq:br_token}
\end{equation}
where $\mathbf{h}^{\mathrm{var}}_{j_t}(n_t)$ is the local embedding of the selected variable.

\paragraph{Online availability and causal construction.}
Seq-BnB is constructed online and uses only information that is available when a solver decision is scored. At a node-selection decision, the solver has an open-node set $\mathcal O_t$, the incumbent information available at that time, global bounds, and the stored attributes of open nodes. Seq-BnB scores each open node using its node token in the current sequence. It does not solve LP relaxations for unprocessed open nodes only for the purpose of node selection. The local LP graph representation is computed only after the selected node $n_t$ is processed and its LP relaxation is solved. Therefore, the summary $\mathbf{s}^{\mathrm{node}}(n_t)$ in the node-selection action token in Eq. \eqref{eq:ns_token} is a record of a completed selection event and is not used to score $n_t$ itself. Similarly, the variable embedding $\mathbf{h}^{\mathrm{var}}_{j_t}(n_t)$ in the branching action token in Eq.  \eqref{eq:br_token} is available only after the LP relaxation at $n_t$ has been solved and the branching candidate set has been formed.

Equivalently, the sequence is causal, i.e., before a decision is scored, it contains only tokens whose corresponding solver events have already occurred. The token recording the current node-selection decision is appended only after that node has been selected, and the token recording the current branching decision is appended only after the branching variable has been selected. These action tokens may condition future decisions, but they cannot affect the scores that produced the actions themselves. The global gap feature $g_t$ is also computed from the incumbent and global bound available at the current decision time; before the first incumbent is found, a fixed missing-value encoding is used. Table~\ref{tab:information_availability} in Appendix~\ref{app:token_details} summarizes the information used by each token type.

Each token is embedded by combining a type embedding and a type-specific feature projection:
\begin{equation}
\mathbf{e}_i=\mathbf{e}_{\mathrm{type}}(\kappa(z_i))+
\mathrm{MLP}_{\kappa(z_i)}(\mathbf{x}(z_i)) \in \mathbb{R}^{d},
\label{eq:token_embedding}
\end{equation}
where $\kappa(z_i)$ is the token type. To control memory and attention cost, we keep a recent history window of at most $L_{\max}$ tokens and force the tokens associated with currently open nodes to remain available. This choice keeps recent search history while preserving the candidates needed for node selection.

\subsection{Global Sequence Encoding and Local Graph Encoding}\label{sec:encoding}
The embedded token matrix is
\begin{equation}
\mathbf{E}_t=[\mathbf{e}_1,\ldots,\mathbf{e}_{L_t}]^\top\in\mathbb{R}^{L_t\times d}.
\end{equation}
We pass $\mathbf{E}_t$ through a self-attention encoder to obtain contextual search-state representations:
\begin{equation}
\mathbf{H}_t=\mathrm{SelfAttention}(\mathbf{E}_t,M_{\mathrm{pad}})\in\mathbb{R}^{L_t\times d},
\label{eq:self_attn}
\end{equation}
where $M_{\mathrm{pad}}$ masks invalid positions. The node-order feature $s(u)$ and action-step feature $t$ provide explicit temporal information to the encoder, allowing it to relate current decisions to earlier tree events.

For the current decision node $n_t$, we encode the LP relaxation as a bipartite graph
\begin{equation}
\mathcal{G}(n_t)=(\mathcal{C}(n_t),\mathcal{V}(n_t),\mathcal{E}(n_t)),
\end{equation}
where constraint nodes and variable nodes are connected by nonzero constraint coefficients. A graph neural network produces variable embeddings
\begin{equation}
\mathbf{H}^{\mathrm{var}}(n_t)=\mathrm{GNN}(\mathcal{G}(n_t))
\in\mathbb{R}^{|\mathcal{V}(n_t)|\times d}.
\end{equation}
The embedding of candidate variable $x_j$ is denoted by $\mathbf{h}^{\mathrm{var}}_j(n_t)$. We also define a node summary
\begin{equation}
\mathbf{s}^{\mathrm{node}}(n_t)=\mathrm{Pool}(\mathbf{H}^{\mathrm{var}}(n_t)),
\end{equation}
which is used in node-selection tokens.

\subsection{Context-Conditioned Decision Heads}\label{sec:heads}
For branching, Seq-BnB uses each candidate variable embedding as a query and the sequence representation as keys and values:
\begin{equation}
\widetilde{\mathbf{h}}^{\mathrm{var}}_j=
\mathrm{CrossAttention}(\mathbf{h}^{\mathrm{var}}_j(n_t),\mathbf{H}_t).
\label{eq:cross_branch}
\end{equation}
The branching score is
\begin{equation}
\hat{y}^{\mathrm{branch}}_j=
\mathrm{MLP}_{\mathrm{branch}}([\mathbf{h}^{\mathrm{var}}_j(n_t);\widetilde{\mathbf{h}}^{\mathrm{var}}_j]).
\label{eq:branch_score}
\end{equation}
The variable with the largest score is selected for branching.

For node selection, each open node $u\in\mathcal{O}_t$ has a corresponding node token in the sequence. Let $\mathbf{h}^{\mathrm{node}}_u$ be its contextual representation in $\mathbf{H}_t$. Seq-BnB scores open nodes by
\begin{equation}
\hat{y}^{\mathrm{ns}}_u=\mathrm{MLP}_{\mathrm{ns}}(\mathbf{h}^{\mathrm{node}}_u),\qquad u\in\mathcal{O}_t,
\label{eq:ns_score}
\end{equation}
then chooses
\begin{equation}
 n_t^*=\arg\max_{u\in\mathcal{O}_t}\hat{y}^{\mathrm{ns}}_u.
\end{equation}

The two heads share the same sequence encoder. This sharing is central to the formulation: node selection and branching are evaluated against a common representation of the tree generated so far, rather than against two disconnected local states. The representation can capture the current gap, the distribution of open nodes, recent tree-growth patterns, and the timing of incumbent updates. Additional action heads could query the same sequence context when extending Seq-BnB to other solver decisions.

\section{Training}\label{sec:training}
\subsection{DAgger Data Aggregation}\label{sec:dagger}
Imitating expert trajectories alone \citep{pomerleau1991efficient} can lead to distribution shift because learned decisions change later search states. Seq-BnB therefore uses DAgger \citep{ross2011reduction}. Branching labels come from strong branching. For node selection, each training MILP is solved offline to obtain a reference optimum $x^\star$. An open node is compatible if its branching bounds still contain $x^\star$; among compatible nodes, the oracle prefers the deepest node, with deterministic bound-based tie-breaking. If no compatible open node remains, it selects by a deterministic lower-bound rule. The reference solution is used for training labels only, never as an inference feature. Appendix~\ref{app:training_details} gives the exact compatibility and tie-breaking formulas.

Starting from expert trajectories, each DAgger round deploys the current policy in the solver, labels the resulting states with the two experts, aggregates them, and retrains the joint policy. This exposes training to the distribution of states created by learned decisions. Algorithm~\ref{alg:dagger} summarizes the procedure.

\begin{algorithm}[t]
\caption{DAgger training for Seq-BnB}\label{alg:dagger}
\scriptsize
\begin{algorithmic}[1]
\State Collect expert trajectories and initialize $\widetilde{\mathcal{D}}_0\leftarrow\mathcal{D}_0$.
\State Train initial policy $\hat{\pi}_0$ on $\widetilde{\mathcal{D}}_0$.
\For{$k=1,\ldots,K$}
    \State Run the solver with policy $\hat{\pi}_{k-1}$ and collect unlabeled decision states $\mathcal{D}^{\mathrm{raw}}_k$.
    \State Label branching states with strong branching and node-selection states with the oracle.
    \State Add the labeled states to the aggregate data set: $\widetilde{\mathcal{D}}_k\leftarrow\widetilde{\mathcal{D}}_{k-1}\cup\mathcal{D}_k$.
    \State Train $\hat{\pi}_k$ on $\widetilde{\mathcal{D}}_k$.
\EndFor
\State \Return $\hat{\pi}_K$.
\end{algorithmic}
\end{algorithm}

\subsection{Joint Loss}\label{sec:loss}
The two decision heads share the sequence encoder and minimize
\begin{equation}
\mathcal L=\mathcal L_{\mathrm{branch}}+\lambda\mathcal L_{\mathrm{ns}},
\label{eq:joint_loss}
\end{equation}
where each term is a cross-entropy loss over the corresponding candidate set and $\lambda$ controls the relative node-selection weight. The complete losses are in Appendix~\ref{app:training_details}.

\section{Computational Experiments}\label{sec:experiments}
\subsection{Experimental Setup}\label{sec:setup}
We evaluate four standard MILP families: Set Covering \citep{balas2009set}, Combinatorial Auction \citep{leyton2000towards}, Capacitated Facility Location \citep{cornuejols1991comparison}, and Max Independent Set \citep{bergman2016decision,alvarez2017machine}. Each has 100 standard test instances and 100 larger transfer instances; sizes are reported in Appendix~\ref{app:datasets}.

We compare against heuristic and learning-based combinations. We use L2B \citep{gasse2019exact} and L2C \citep{labassi2022learning} as graph-based, imitation-learning single-decision baselines: their local representations and learning settings provide closer reference points for assessing the contribution of Seq-BnB's shared search-history representation. T-BranT \citep{lin2022tbrant} and Mamba-Branching \citep{zhang2026mamba} introduce different history-aware branching architectures, while TRGNN \citep{zhang2025trgnn} uses a tripartite tree representation and reinforcement learning for node selection; we discuss these methods in Section~\ref{sec:related_work} rather than treating them as controlled baselines of our joint architecture. Du et al.'s joint node-and-variable selection method \citep{du2025learning} is closely related, but we could not locate publicly available code or complete a reliable reimplementation, so it is not included in the experimental comparison. Branching baselines include strong branching (SB), pseudocost branching (PSB), the GNN branching model L2B \citep{gasse2019exact}, and Retro \citep{parsonson2023reinforcement}. Node-selection baselines include Oracle, Best Estimate Search, RankNet \citep{song2018learning}, and L2C \citep{labassi2022learning}. The expert SB+Oracle is included as a reference and is not counted when selecting the best nonexpert result.

All methods use SCIP 7.0.3, with cuts only at the root and no restarts as in prior branching studies \citep{karzan2009information,fischetti2012branching,khalil2016learning}, and a 3,600-second limit. Time and processed-node counts are shifted geometric means (shift 1); unsolved instances contribute their final time and processed-node count. Each learning-based method uses 120,000 training samples. Seq-BnB starts with 10,000 expert samples and collects about 10,000 per round over 11 DAgger rounds. We use Adam, batch size 64, learning rate $10^{-5}$, $L_{\max}=500$, and $\lambda=0.1$. Training uses one NVIDIA A100 GPU; inference uses one thread and fixed seeds. The code used for training and evaluation will be made publicly available upon publication.

\subsection{Main Results}\label{sec:main_results}
Table~\ref{tab:main_test} reports the standard test sets. Seq-BnB processes the fewest nodes on all four families. It is fastest on Set Covering and Max Independent Set; on Combinatorial Auction and Facility Location its time remains close to the fastest method. Relative to L2B+Estimate, the strongest learning-based branching combination in these experiments, Seq-BnB uses fewer nodes in every standard family.

Table~\ref{tab:main_transfer} evaluates larger instances from the same families. Seq-BnB again processes the fewest nodes in all four and achieves the fastest time on Facility Location and Max Independent Set. On Set Covering and Combinatorial Auction, PSB+Estimate and L2B+Estimate are faster, respectively. This distinction matters: processed nodes reflect search effort within the time limit, whereas wall-clock time also includes sequence construction, model inference, and varying LP costs. The transfer results show that the search benefit persists as instance size grows, while inference overhead remains a practical target for improvement. Complete results including wins appear in Appendix~\ref{app:full_results}.

\begin{table}[t]
\TABLE
{Standard test results (100 instances per family).\label{tab:main_test}}
{\scriptsize
\begin{tabular}{l*{8}{r}}
\hline
\multirow{2}{*}{Method} & \multicolumn{2}{c}{Set Covering} & \multicolumn{2}{c}{Comb. Auction} & \multicolumn{2}{c}{Facility Location} & \multicolumn{2}{c}{Max Ind. Set}\\
\cline{2-9}
& Time & Nodes & Time & Nodes & Time & Nodes & Time & Nodes\\
\hline\up
SB+Oracle & 9.01 & 20.89 & 59.34 & 183.28 & 82.99 & 131.58 & 101.37 & 30.44\\
\hline
PSB+Estimate & 3.73 & 37.25 & 10.98 & 886.54 & 41.33 & 226.81 & 9.88 & 101.93\\
RankNet+PSB & 3.87 & 37.49 & 16.52 & 1247.04 & 28.70 & 256.24 & 11.42 & 86.91\\
L2C+PSB & 3.60 & 38.86 & 29.75 & 1806.76 & 30.80 & 265.38 & 7.78 & 156.95\\
L2B+Estimate & 2.99 & 32.06 & \textbf{10.13} & 292.23 & \textbf{28.60} & 180.19 & 6.45 & 38.10\\
Retro+Estimate & 6.36 & 136.16 & 18.44 & 1124.61 & 35.36 & 230.98 & 12.27 & 144.39\\
\hline
\textbf{Seq-BnB} & \textbf{2.81} & \textbf{28.62} & 10.61 & \textbf{252.34} & 30.22 & \textbf{164.16} & \textbf{6.35} & \textbf{36.54}\\ \hline
\end{tabular}}
{Time is in seconds. Time and Nodes are shifted geometric means; bold marks the best nonexpert method. Full results including wins are in Appendix~\ref{app:full_results}.}
\end{table}

\begin{table}[t]
\TABLE
{Results on larger transfer instances (100 instances per family).\label{tab:main_transfer}}
{\scriptsize
\begin{tabular}{l*{8}{r}}
\hline
\multirow{2}{*}{Method} & \multicolumn{2}{c}{Set Covering} & \multicolumn{2}{c}{Comb. Auction} & \multicolumn{2}{c}{Facility Location} & \multicolumn{2}{c}{Max Ind. Set}\\
\cline{2-9}
& Time & Nodes & Time & Nodes & Time & Nodes & Time & Nodes\\
\hline\up
SB+Oracle & 121.58 & 234.79 & 3363.43 & 2074.32 & 2304.62 & 207.62 & 2795.54 & 183.65\\
\hline
PSB+Estimate & \textbf{25.66} & 667.39 & 2174.09 & 21105.59 & 748.19 & 1390.19 & 1217.84 & 13517.28\\
RankNet+PSB & 40.38 & 802.18 & 1228.58 & 37844.01 & 2113.60 & 4398.48 & 644.66 & 11997.45\\
L2C+PSB & 45.72 & 696.74 & 1526.97 & 66897.30 & 1407.08 & 3611.90 & 601.47 & 16559.32\\
L2B+Estimate & 28.97 & 462.74 & \textbf{618.79} & 11771.63 & 747.34 & 1475.62 & 459.58 & 6085.88\\
Retro+Estimate & 514.17 & 7766.10 & 3591.17 & 27588.27 & 1162.38 & 1740.44 & 2171.13 & 18567.67\\
\hline
\textbf{Seq-BnB} & 37.33 & \textbf{402.63} & 790.99 & \textbf{5134.41} & \textbf{621.68} & \textbf{1249.23} & \textbf{412.42} & \textbf{1179.97}\\ \hline
\end{tabular}}
{Time is in seconds. Time and Nodes are shifted geometric means; bold marks the best nonexpert method. Full results including wins are in Appendix~\ref{app:full_results}.}
\end{table}

\subsection{Ablations}\label{sec:ablation}
Table~\ref{tab:ablation_summary} tests the central design choices on standard instances. Removing sequence context from the branching head increases both time and processed nodes in every family, showing the value of history-conditioned branching. The BR-only variant pairs learned branching with Estimate, and NS-only pairs learned node selection with PSB. The integrated policy outperforms both, supporting the value of learning the two coupled decisions together. Training without DAgger uses static expert data of the same total sample size and performs worse throughout, consistent with the policy-induced state-distribution motivation.

Appendix~\ref{app:ablation_details} gives the complete tables and branching imitation accuracy. Appendix~\ref{app:sensitivity} analyzes $L_{\max}$ and $\lambda$: the default settings attain the best time and processed-node count among the tested values on all four families.

\begin{table}[t]
\TABLE
{Core ablations on standard test instances (Time/Nodes).\label{tab:ablation_summary}}
{\scriptsize
\begin{tabular}{l*{8}{r}}
\hline
\multirow{2}{*}{Method} & \multicolumn{2}{c}{Set Covering} & \multicolumn{2}{c}{Comb. Auction} & \multicolumn{2}{c}{Facility Location} & \multicolumn{2}{c}{Max Ind. Set}\\
\cline{2-9}
& Time & Nodes & Time & Nodes & Time & Nodes & Time & Nodes\\
\hline\up
w/o Seq (branch head) & 3.51 & 31.94 & 11.83 & 281.63 & 41.56 & 238.42 & 7.48 & 42.73\\
BR-only & 3.04 & 34.24 & 12.18 & 306.54 & 39.56 & 225.63 & 6.92 & 39.84\\
NS-only & 3.80 & 32.25 & 27.47 & 684.73 & 40.88 & 236.94 & 7.89 & 68.74\\
w/o DAgger & 5.36 & 43.21 & 18.21 & 397.63 & 49.37 & 374.35 & 8.64 & 49.51\\
\hline
\textbf{Seq-BnB} & \textbf{2.81} & \textbf{28.62} & \textbf{10.61} & \textbf{252.34} & \textbf{30.22} & \textbf{164.16} & \textbf{6.35} & \textbf{36.54}\\ \hline
\end{tabular}}
{Time is in seconds; Time and Nodes are shifted geometric means. Detailed tables and branching accuracy are in Appendix~\ref{app:ablation_details}.}
\end{table}

\section{Conclusion}\label{sec:conclusion}
This paper presents Seq-BnB, a sequence-structured framework for jointly learning branching and node selection in MILP branch-and-bound. These decisions are coupled because node selection determines where search continues and branching determines how that region expands. Seq-BnB represents their common consequence, the evolving search tree and its decision trajectory, through typed tokens processed by a shared self-attention encoder. A bipartite GNN supplies local LP structure, cross-attention connects global context to local candidates, and DAgger-style data aggregation reduces the distribution shift induced by deploying learned policies inside the solver.

Across four MILP benchmark families, Seq-BnB consistently reduces the number of processed BnB nodes and often improves solution time. The joint-learning ablation supports the value of modeling branching and node selection together, while transfer experiments show that the sequence context remains useful on larger instances. Beyond the two actions evaluated here, the sequence representation provides a natural interface for integrating additional solver decisions, such as cut selection or primal-heuristic scheduling, through new token types and decision heads. Evaluating such extensions and reducing runtime overhead through cached token embeddings, batched candidate scores, or lighter sequence encoders are directions for future work.

\Needspace{12\baselineskip}
\section*{Acknowledgments}
This work is supported by the National Natural Science Foundation of China (NSFC) under Grant No.~72401122, Guangdong Basic and Applied Basic Research Foundation under Grants No.~2024A1515012241 and 2021A1515110024, Guangdong University Young Innovative Talents Program Project under Grant No.~2024KQNCX050, Shenzhen Fundamental Research Program under Grant No.~JCYJ20200109141235597, and the NSFC under Grant No.~61761136008.

\subsection*{AI use statement}
During the preparation of this work, the authors used ChatGPT (OpenAI) to assist with
language refinement, readability improvement, and the organization of manuscript content.
After using this tool, the authors carefully reviewed and edited the manuscript as needed
and take full responsibility for the content of the published article.

\bibliographystyle{plainnat}
\bibliography{references}
\clearpage
\appendix
\section{Additional Background and Related Work}\label{app:related_work}
\subsection{MILP}
This paper considers a MILP of the form
\begin{equation}
\min \{ c^\top x : Ax \le b,\ x_j \in \mathbb{Z}\ \text{for}\ j\in \mathcal{I} \},
\label{eq:milp}
\end{equation}
where $x$ contains continuous and integer variables. BnB solves \eqref{eq:milp} by repeatedly solving LP relaxations over subproblems. If the LP solution at a node is infeasible, integer feasible, or cannot improve the incumbent, the node is fathomed. Otherwise, the solver branches on a fractional integer variable and creates child nodes with additional bounds or disjunctive constraints.

At each search step $t$, the solver maintains an open-node set $\mathcal{O}_t$, an incumbent solution if one has been found, and global bounds. Two tree-construction decisions are made repeatedly. The node-selection policy chooses the next node $n_t\in\mathcal{O}_t$ to process, thereby choosing the current region of exploration. The branching policy then chooses a candidate fractional variable $x_{j_t}$ at $n_t$ and creates child subproblems, thereby determining the local expansion of that region. Because each branching action changes the future open-node set and each node-selection action determines where subsequent branching is applied, the two policies jointly induce the evolving search tree. 

% We view these decisions as actions in a finite-horizon Markov decision process \citep{sutton1998reinforcement}:
% \begin{equation}
% \mathcal{M}=(\mathcal{S},\mathcal{A},P,r,\gamma),
% \end{equation}
% where the state records the current BnB search status, the action includes node selection and branching, and the transition is induced by the solver after applying the action. This formulation is not used to optimize a reward directly. It motivates why search history and policy-induced distribution shift matter for imitation learning in BnB.

\subsection{BnB Search and Solver Decisions}
BnB is the backbone of exact MILP solving. Its classical form recursively partitions the feasible region and uses relaxations to prune subproblems that cannot improve the incumbent solution \citep{land1960automatic,wolsey1998integer}. Modern MILP solvers implement a richer branch-and-cut framework, where presolving, cutting planes, primal heuristics, conflict analysis, and tree-search rules interact throughout the solution process \citep{achterberg2009scip,gamrath2020scip}. 

Among these components, branching and node selection are especially central because they determine how the BnB tree is generated and explored. Classical branching rules include strong branching, pseudocost branching, and reliability branching, while node-selection rules include depth-first, best-bound, best-estimate, and hybrid search strategies \citep{achterberg2005branching,linderoth1999computational}. Prior computational studies show that these rules can have a large impact on both tree size and solution time. Our work follows this solver-centered perspective, but treats branching and node selection as coupled decisions over a shared evolving tree rather than as independent heuristic modules.

\subsection{Learning Branching and Node Selection Policies}
Learning-based MILP research has mainly focused on replacing or approximating individual BnB decisions, reflecting a broader trend of integrating machine learning with combinatorial optimization solvers \citep{lodi2017learning,bengio2021machine}. For branching, early approaches learn variable scores or rankings from strong-branching demonstrations using hand-crafted features \citep{khalil2016learning,alvarez2017machine}. Later work represents the LP relaxation as a variable--constraint bipartite graph and uses graph neural networks to imitate strong branching \citep{gasse2019exact}. Several studies further incorporate search-tree state or trajectory information into branching, for example by parameterizing the BnB tree state \citep{zarpellon2021parameterizing}, exploiting parent--child dependencies in branching labels \citep{gupta2022lookback}, or learning branching through reinforcement learning and retrospective trajectories \citep{etheve2020reinforcement,parsonson2023reinforcement}. More recently, T-BranT integrates branching history through a tree-aware transformer \citep{lin2022tbrant}, while Mamba-Branching models sequences of branching states and decisions \citep{zhang2026mamba}.

For node selection, \citet{he2014learning} apply imitation learning to learn adaptive BnB search orders, while \citet{labassi2022learning} use a siamese graph neural network to compare open nodes. TRGNN represents the evolving BnB tree as a tripartite graph for node selection \citep{zhang2025trgnn}. These methods provide strong evidence that BnB decisions can be learned from data. However, they typically optimize either branching or node selection in isolation. Seq-BnB is complementary to these works: it uses a bipartite graph encoder for local LP structure, but conditions both branching and node-selection heads on the same global sequence representation.

% \subsection{Learning Other Solver Components and Global Search Context}
% Beyond branching and node selection, recent work has applied learning to other solver components. Neural diving and related primal-learning approaches predict high-quality assignments or partial assignments to accelerate feasible-solution discovery \citep{nair2020solving,sonnerat2021learning}. Learning-based heuristic scheduling aims to decide when and how to call primal heuristics inside BnB \citep{chmiela2021learning}. Cut selection has also attracted increasing attention, with imitation learning, reinforcement learning, and survey work showing that the choice of cuts can be treated as a learnable solver decision \citep{paulus2022learning,turner2023adaptive,deza2023machine}. These studies suggest that a general-purpose learning interface for BnB should not be limited to a single action type. Existing methods, however, often build task-specific representations and heads for one solver component. Seq-BnB instead represents the evolving search process as a typed sequence of node and action tokens. This design allows branching and node selection to share global context in the present paper, and it also provides a natural path for incorporating additional solver actions such as cut selection or heuristic scheduling.

\subsection{Learning Other Solver Components and Global Search Context}
Beyond branching and node selection, recent work has applied learning to other solver components. Neural diving and related primal-learning approaches predict high-quality assignments or partial assignments to accelerate feasible-solution discovery \citep{nair2020solving,sonnerat2021learning}. Learning-based heuristic scheduling aims to decide when and how to call primal heuristics inside BnB \citep{chmiela2021learning}. Cut selection has also attracted increasing attention, with imitation learning, reinforcement learning, and survey work showing that the choice of cuts can be treated as a learnable solver decision \citep{paulus2022learning,turner2023adaptive,deza2023machine}.

Recent studies have further moved beyond isolated solver decisions by exploring shared representations across tasks or coupled policies within BnB. Cai et al.~\citep{cai2024multi} proposed a multi-task representation learning framework that learns shared MILP embeddings across backdoor identification, predict-and-search, and solver configuration, while using task-specific output heads for different downstream decisions. From a broader generalization perspective, Li et al.~\citep{li2024towards} investigated a foundation-model paradigm for MILP, training a single model on a large and diverse collection of generated MILP classes for integrality-gap prediction, learning to branch, and language--MILP alignment. More closely related to the coupling of BnB decisions, Du et al.~\citep{du2025learning} developed a unified learning-based framework that jointly integrates node selection and variable selection, highlighting the interdependence between these two decisions during the search process.

These works suggest that a general learning interface for MILP solving should not be restricted to a single solver component. Nevertheless, existing multi-task and foundation-model approaches primarily focus on representations shared across tasks or problem classes; history-aware branching and tree-aware node-selection methods address one decision type at a time, while prior joint BnB policies do not study a shared chronological sequence of both node-selection and branching events. In contrast, Seq-BnB represents an evolving BnB run as a typed sequence of node and action tokens and combines this search-history representation with the current MILP bipartite graph. This design enables branching and node selection to condition on a shared global search context, rather than relying only on local solver states or static instance-level embeddings. Although the present work focuses on branching and node selection, the same sequence-based interface could be extended to incorporate additional solver actions, such as cut selection, heuristic scheduling, or other learned control decisions.

\section{Token Features and Online Information}\label{app:token_details}
Table~\ref{tab:information_availability} summarizes when each feature group becomes available and how it is used.

\begin{table}[t]
\caption{Information availability in the online construction of Seq-BnB.}
\label{tab:information_availability}
\scriptsize
\setlength{\tabcolsep}{3pt}
\begin{tabular}{p{0.16\linewidth}p{0.3\linewidth}p{0.3\linewidth}p{0.16\linewidth}}
\hline\up
Token or Quantity & Feature Group & Availability & Use in Scoring \\
\hline\up
Open-node token 
& Stored node bound $\ell(u)$, estimate $c(u)$, depth $d(u)$, creation/visit order $s(u)$, and currently available global statistics 
& Available before node selection from the solver's open-node data structure; no LP relaxation is solved solely for scoring open nodes 
& Scores node selection through Eq. \eqref{eq:ns_score} \\
Processed-node update 
& Visited, fathomed, and incumbent-update indicators, together with realized node status 
& Available only after the corresponding node has been processed or fathomed, or after an incumbent update has occurred 
& Past context only \\
Node-selection action token 
& Selected-node statistics $d(n_t)$, $\ell(n_t)$, $c(n_t)$, and the local summary $\mathbf{s}^{\mathrm{node}}(n_t)$ after the selected node is processed 
& Created after the node-selection decision; $\mathbf{s}^{\mathrm{node}}(n_t)$ is not available before selecting $n_t$ 
& Not used to choose $n_t$; used only for branching at the processed node and future decisions \\
Branching candidates 
& Local variable embeddings $\mathbf{h}^{\mathrm{var}}_j(n_t)$ for fractional candidate variables 
& Available after solving the LP relaxation at the selected node $n_t$ 
& Scores branching through Eqs. \eqref{eq:cross_branch} and \eqref{eq:branch_score}\\
Branching action token 
& Selected-variable embedding $\mathbf{h}^{\mathrm{var}}_{j_t}(n_t)$ and current search statistics 
& Created after the branching variable has been selected 
& Future context only \\
\hline\up
\end{tabular}
\end{table}

\section{Expert Oracle and Training Details}\label{app:training_details}
\paragraph{Oracle labels for node selection.}
For each training instance, we first solve the MILP to optimality offline and fix one reference optimal solution $x^\star$. 
% If multiple optimal solutions are returned by the solver, we use the first optimal solution obtained under the fixed random seed and solver settings. 
The reference solution is used only for producing training labels and is never provided to Seq-BnB at test time.

Each BnB node $u$ defines a local domain through the branching decisions on the path from the root to $u$. 
Let $L_j(u)$ and $U_j(u)$ denote the lower and upper bounds imposed on integer variable $x_j$ at node $u$, including both the original bounds and all branching-bound changes along the path. 
We say that node $u$ is compatible with the reference solution $x^\star$ if
\[
    L_j(u) \le x^\star_j \le U_j(u), \qquad \forall j \in \mathcal I .
\]
Equivalently, $x^\star$ satisfies all branching decisions that define node $u$. 
At node-selection state $t$, define the compatible open-node set
\[
    \mathcal C_t(x^\star) =
    \{u \in \mathcal O_t : L_j(u) \le x^\star_j \le U_j(u),\ \forall j \in \mathcal I \}.
\]
If $\mathcal C_t(x^\star)$ is nonempty, the oracle selects a compatible node according to a deterministic tie-breaking rule:
\[
    u_t^\star
    =
    \operatorname{lexmin}_{u \in \mathcal C_t(x^\star)}
    \bigl(-d(u),\ \ell(u),\ c(u),\ s(u)\bigr).
\]
Thus, among all open nodes that can still contain the reference optimum, the oracle first prefers the deepest node, then the smaller LP lower bound, then the smaller estimate value, and finally the earlier creation order. 
This rule follows a path toward the fixed optimal solution while keeping the label deterministic when several compatible nodes exist.

If $\mathcal C_t(x^\star)$ is empty, which can occur after the reference-optimal path has already been processed or when a learner-induced state no longer contains a compatible open node, the oracle falls back to a deterministic proof-oriented rule:
\[
    u_t^\star
    =
    \operatorname{lexmin}_{u \in \mathcal O_t}
    \bigl(\ell(u),\ c(u),\ -d(u),\ s(u)\bigr).
\]
That is, it selects the open node with the smallest lower bound, breaking ties by estimate value, depth, and creation order. 
The compatibility test only uses the branching bounds that define each open node and the fixed reference solution $x^\star$; it does not require future node outcomes or LP information from unprocessed open nodes. 
Therefore, the oracle label is well-defined not only on expert-generated trees, but also on learner-induced trees encountered during DAgger.

\subsection{Candidate-Set Losses}
For a branching sample $i$, let $\mathcal{V}_i$ be the candidate set and let $a_i^*$ be the expert variable. The branching loss is
\begin{equation}
\mathcal{L}_{\mathrm{branch}}=-\frac{1}{|\mathcal{B}|}
\sum_{i\in\mathcal{B}}
\log\frac{\exp(\hat{y}^{\mathrm{branch}}_{a_i^*})}
{\sum_{j\in\mathcal{V}_i}\exp(\hat{y}^{\mathrm{branch}}_j)}.
\label{eq:branch_loss}
\end{equation}

For a node-selection sample $i$, let $\mathcal{O}_i$ be the open-node set and $u_i^\star$ be the oracle node defined by the compatibility rule in Section~\ref{sec:dagger}. The node-selection loss is
\begin{equation}
\mathcal{L}_{\mathrm{ns}}=-\frac{1}{|\mathcal{I}_{\mathrm{ns}}|}
\sum_{i\in\mathcal{I}_{\mathrm{ns}}}
\log\frac{\exp(\hat{y}^{\mathrm{ns}}_{u_i^*})}
{\sum_{u\in\mathcal{O}_i}\exp(\hat{y}^{\mathrm{ns}}_u)}.
\label{eq:ns_loss}
\end{equation}
This loss is aligned with inference, where the solver selects the maximum-scoring candidate from the relevant candidate set.

\section{Instance Statistics}\label{app:datasets}
Table~\ref{tab:dataset_stats} lists the instance sizes and sample counts for the standard and larger transfer sets used in Section~\ref{sec:main_results}.

\begin{table}[t]
\TABLE
{Problem instance statistics for test and transfer sets.\label{tab:dataset_stats}}
{\scriptsize
\begin{tabular}{l*{8}{r}}
\hline\up
\multirow{2}{*}{Statistic} & \multicolumn{2}{c}{Set Covering} & \multicolumn{2}{c}{Comb. Auction} & \multicolumn{2}{c}{Facility Location} & \multicolumn{2}{c}{Max Ind. Set}\\
\cline{2-9}
& Test & Transfer & Test & Transfer & Test & Transfer & Test & Transfer\\
\hline\up
Avg. binary variables & 1000.00 & 1000.00 & 500.00 & 1000.00 & 100.00 & 200.00 & 500.00 & 1000.00\\
Avg. integer variables & 0.00 & 0.00 & 0.00 & 0.00 & 0.00 & 0.00 & 0.00 & 0.00\\
Avg. continuous variables & 0.00 & 0.00 & 0.00 & 0.00 & 10000.00 & 40000.00 & 0.00 & 0.00\\
Avg. constraints & 500.00 & 1000.00 & 191.90 & 386.10 & 10201.00 & 40401.00 & 2088.60 & 4219.30\\
Instances & 100 & 100 & 100 & 100 & 100 & 100 & 100 & 100\\ \hline
\end{tabular}}
{All values are averages over the corresponding set.}
\end{table}

\section{Complete Main Results}\label{app:full_results}
Tables~\ref{tab:main_test_full} and~\ref{tab:main_transfer_full} provide the full standard-test and transfer results underlying Section~\ref{sec:main_results}, including the win counts omitted from the compact main-text tables. A win records the shortest solving time on an instance among nonexpert methods; SB+Oracle is an expert reference.

\begin{table}[t]
\TABLE
{Results on standard test instances.\label{tab:main_test_full}}
{\scriptsize
\begin{tabular}{l*{12}{r}}
\hline
\multirow{2}{*}{Method} & \multicolumn{3}{c}{Set Covering} & \multicolumn{3}{c}{Comb. Auction} & \multicolumn{3}{c}{Facility Location} & \multicolumn{3}{c}{Max Ind. Set}\\
\cline{2-13}
& Time & Nodes & Wins & Time & Nodes & Wins & Time & Nodes & Wins & Time & Nodes & Wins\\
\hline\up
SB+Oracle & 9.01 & 20.89 & -- & 59.34 & 183.28 & -- & 82.99 & 131.58 & -- & 101.37 & 30.44 & --\\
\hline
PSB+Estimate & 3.73 & 37.25 & 8 & 10.98 & 886.54 & 9 & 41.33 & 226.81 & 1 & 9.88 & 101.93 & 6\\
RankNet+PSB & 3.87 & 37.49 & 12 & 16.52 & 1247.04 & 1 & 28.70 & 256.24 & 28 & 11.42 & 86.91 & 4\\
L2C+PSB & 3.60 & 38.86 & 12 & 29.75 & 1806.76 & 0 & 30.80 & 265.38 & 21 & 7.78 & 156.95 & 15\\
L2B+Estimate & 2.99 & 32.06 & 24 & \textbf{10.13} & 292.23 & \textbf{49} & \textbf{28.60} & 180.19 & \textbf{30} & 6.45 & 38.10 & 27\\
Retro+Estimate & 6.36 & 136.16 & 0 & 18.44 & 1124.61 & 0 & 35.36 & 230.98 & 0 & 12.27 & 144.39 & 0\\
\hline
\textbf{Seq-BnB} & \textbf{2.81} & \textbf{28.62} & \textbf{44} & 10.61 & \textbf{252.34} & 41 & 30.22 & \textbf{164.16} & 20 & \textbf{6.35} & \textbf{36.54} & \textbf{48}\\ \hline
\end{tabular}}
{Time is in seconds. Time and node counts are shifted geometric means. Bold values mark the best nonexpert method.}
\end{table}

\begin{table}[t]
\TABLE
{Results on transfer instances.\label{tab:main_transfer_full}}
{\scriptsize
\begin{tabular}{l*{12}{r}}
\hline
\multirow{2}{*}{Method} & \multicolumn{3}{c}{Set Covering} & \multicolumn{3}{c}{Comb. Auction} & \multicolumn{3}{c}{Facility Location} & \multicolumn{3}{c}{Max Ind. Set}\\
\cline{2-13}
& Time & Nodes & Wins & Time & Nodes & Wins & Time & Nodes & Wins & Time & Nodes & Wins\\
\hline
SB+Oracle & 121.58 & 234.79 & -- & 3363.43 & 2074.32 & -- & 2304.62 & 207.62 & -- & 2795.54 & 183.65 & --\\
\hline
PSB+Estimate & \textbf{25.66} & 667.39 & \textbf{43} & 2174.09 & 21105.59 & 0 & 748.19 & 1390.19 & 14 & 1217.84 & 13517.28 & 1\\
RankNet+PSB & 40.38 & 802.18 & 1 & 1228.58 & 37844.01 & 1 & 2113.60 & 4398.48 & 0 & 644.66 & 11997.45 & 4\\
L2C+PSB & 45.72 & 696.74 & 2 & 1526.97 & 66897.30 & 0 & 1407.08 & 3611.90 & 0 & 601.47 & 16559.32 & 5\\
L2B+Estimate & 28.97 & 462.74 & 31 & \textbf{618.79} & 11771.63 & \textbf{72} & 747.34 & 1475.62 & 16 & 459.58 & 6085.88 & 17\\
Retro+Estimate & 514.17 & 7766.10 & 0 & 3591.17 & 27588.27 & 0 & 1162.38 & 1740.44 & 1 & 2171.13 & 18567.67 & 0\\
\hline
\textbf{Seq-BnB} & 37.33 & \textbf{402.63} & 23 & 790.99 & \textbf{5134.41} & 27 & \textbf{621.68} & \textbf{1249.23} & \textbf{69} & \textbf{412.42} & \textbf{1179.97} & \textbf{73}\\ \hline
\end{tabular}}
{Time is in seconds. Time and node counts are shifted geometric means. Bold values mark the best nonexpert method.}
\end{table}

\section{Detailed Ablations}\label{app:ablation_details}
\begin{figure}
\FIGURE
{\includegraphics[width=0.82\textwidth]{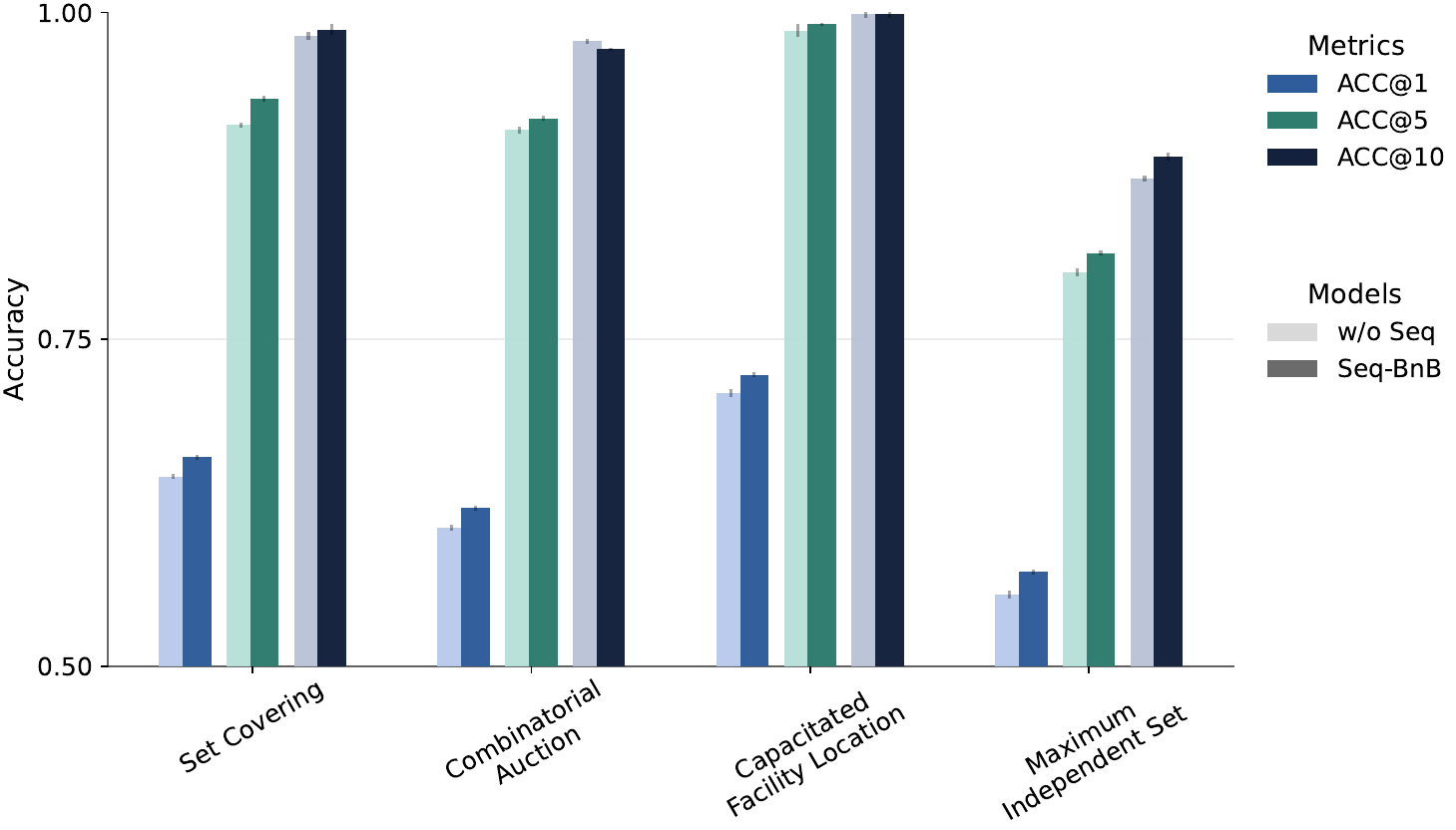}}
{Branching imitation accuracy with and without sequence context. ACC@$k$ measures whether the expert variable is among the top $k$ predicted candidates. Sequence context improves top-$k$ accuracy on most benchmarks; together with Table~\ref{tab:ablation_seq}, this suggests that modest gains in single-step imitation can translate into larger improvements over the subsequent BnB search.\label{fig:acc}}
{}
\end{figure}

Tables~\ref{tab:ablation_seq}, \ref{tab:ablation_joint}, and~\ref{tab:ablation_dagger} give the full results for the sequence-context, joint-learning, and DAgger ablations summarized in Section~\ref{sec:ablation}. In the w/o Seq variant, the branching head uses only the local variable embedding $\mathbf{h}^{\mathrm{var}}_j(n_t)$. Figure~\ref{fig:acc} shows branching imitation accuracy with and without sequence context.

\begin{table}[t]
\TABLE
{Effect of sequence context on standard test instances.\label{tab:ablation_seq}}
{\scriptsize
\begin{tabular}{l*{12}{r}}
\hline
\multirow{2}{*}{Method} & \multicolumn{3}{c}{Set Covering} & \multicolumn{3}{c}{Comb. Auction} & \multicolumn{3}{c}{Facility Location} & \multicolumn{3}{c}{Max Ind. Set}\\
\cline{2-13}
& Time & Nodes & Wins & Time & Nodes & Wins & Time & Nodes & Wins & Time & Nodes & Wins\\
\hline
Seq-BnB (w/o Seq) & 3.51 & 31.94 & 18 & 11.83 & 281.63 & 21 & 41.56 & 238.42 & 15 & 7.48 & 42.73 & 7\\
\textbf{Seq-BnB} & \textbf{2.81} & \textbf{28.62} & \textbf{82} & \textbf{10.61} & \textbf{252.34} & \textbf{79} & \textbf{30.22} & \textbf{164.16} & \textbf{85} & \textbf{6.35} & \textbf{36.54} & \textbf{93}\\ \hline
\end{tabular}}
{Wins are counted between the two compared methods.}
\end{table}

\begin{table}[t]
\TABLE
{Joint learning ablation\label{tab:ablation_joint}}
{\scriptsize
\begin{tabular}{l*{12}{r}}
\hline
\multirow{2}{*}{Method} & \multicolumn{3}{c}{Set Covering} & \multicolumn{3}{c}{Comb. Auction} & \multicolumn{3}{c}{Facility Location} & \multicolumn{3}{c}{Max Ind. Set}\\
\cline{2-13}
& Time & Nodes & Wins & Time & Nodes & Wins & Time & Nodes & Wins & Time & Nodes & Wins\\
\hline
Seq-BnB (BR-only) & 3.04 & 34.24 & 12 & 12.18 & 306.54 & 11 & 39.56 & 225.63 & 3 & 6.92 & 39.84 & 15\\
Seq-BnB (NS-only) & 3.80 & 32.25 & 6 & 27.47 & 684.73 & 3 & 40.88 & 236.94 & 5 & 7.89 & 68.74 & 10\\
% TreeCL & 3.01 & 34.51 & 20 & 24.61 & 831.34 & 6 & \textbf{26.22} & 258.28 & \textbf{49} & 7.05 & 80.47 & 6\\
\hline
\textbf{Seq-BnB} & \textbf{2.81} & \textbf{28.62} & \textbf{82} & \textbf{10.61} & \textbf{252.34} & \textbf{86} & \textbf{30.22} & \textbf{164.16} & \textbf{92} & \textbf{6.35} & \textbf{36.54} & \textbf{75}\\ \hline
\end{tabular}}
{}
\end{table}

\begin{table}[t]
\TABLE
{Effect of DAgger data aggregation.\label{tab:ablation_dagger}}
{\scriptsize
\begin{tabular}{l*{12}{r}}
\hline
\multirow{2}{*}{Method} & \multicolumn{3}{c}{Set Covering} & \multicolumn{3}{c}{Comb. Auction} & \multicolumn{3}{c}{Facility Location} & \multicolumn{3}{c}{Max Ind. Set}\\
\cline{2-13}
& Time & Nodes & Wins & Time & Nodes & Wins & Time & Nodes & Wins & Time & Nodes & Wins\\
\hline
Seq-BnB (w/o DAgger) & 5.36 & 43.21 & 5 & 18.21 & 397.63 & 0 & 49.37 & 374.35 & 2 & 8.64 & 49.51 & 4\\
\textbf{Seq-BnB} & \textbf{2.81} & \textbf{28.62} & \textbf{95} & \textbf{10.61} & \textbf{252.34} & \textbf{100} & \textbf{30.22} & \textbf{164.16} & \textbf{98} & \textbf{6.35} & \textbf{36.54} & \textbf{96}\\ \hline
\end{tabular}}
{Wins are counted between the two compared methods.}
\end{table}

\section{Hyperparameter Sensitivity}\label{app:sensitivity}
\subsection{Sequence Length and Loss Coupling}\label{sec:sensitivity}
We further examine two hyperparameters that are directly tied to the proposed formulation: the maximum sequence window length $L_{\max}$ and the joint-loss weight $\lambda$. Unlike generic training hyperparameters, these two quantities have clear algorithmic meanings. The window length $L_{\max}$ controls how much recent BnB history is retained by the global sequence encoder, while $\lambda$ controls the relative strength of node-selection supervision in the shared representation.

Table~\ref{tab:ablation_lmax} reports the effect of varying $L_{\max}$. A larger window provides the self-attention encoder with more search-history information, but also increases memory usage and attention cost. The results show a consistent improvement as $L_{\max}$ increases from 100 to 500. Across all four benchmark families, $L_{\max}=500$ gives the lowest shifted geometric mean in both solving time and processed nodes, and it wins the largest number of instances among the three tested values. This supports the importance of retaining sufficient global search context: Seq-BnB benefits not only from the currently open nodes, but also from recent tree-growth and decision-history information.

\begin{table}[t]
\TABLE
{Effect of maximum sequence-window length $L_{\max}$ on standard test instances.\label{tab:ablation_lmax}}
{\scriptsize
\begin{tabular}{l*{12}{r}}
\hline
\multirow{2}{*}{$L_{\max}$} & \multicolumn{3}{c}{Set Covering} & \multicolumn{3}{c}{Comb. Auction} & \multicolumn{3}{c}{Facility Location} & \multicolumn{3}{c}{Max Ind. Set}\\
\cline{2-13}
& Time & Nodes & Wins & Time & Nodes & Wins & Time & Nodes & Wins & Time & Nodes & Wins\\
\hline
100 & 3.46 & 34.87 & 5 & 13.28 & 334.52 & 1 & 39.84 & 232.65 & 0 & 7.58 & 45.37 & 0\\
300 & 3.07 & 31.46 & 8 & 11.76 & 287.91 & 4 & 34.92 & 193.74 & 6 & 6.91 & 39.82 & 2\\
500 & \textbf{2.81} & \textbf{28.62} & \textbf{87} & \textbf{10.61} & \textbf{252.34} & \textbf{95} & \textbf{30.22} & \textbf{164.16} & \textbf{94} & \textbf{6.35} & \textbf{36.54} & \textbf{98}\\ \hline
\end{tabular}}
{Wins are counted among the three window sizes.}
\end{table}

Table~\ref{tab:ablation_lambda} reports the effect of the joint-loss weight $\lambda$. When $\lambda$ is too small, the shared encoder receives weak supervision from the node-selection task; when $\lambda$ is too large, the node-selection loss can dominate the branching objective and bias the shared representation. The results exhibit this tradeoff clearly. Intermediate values perform better than the two extremes, and the default value $\lambda=0.1$ achieves the best time and node count on all four benchmark families. These results indicate that joint learning is most effective when branching and node selection are balanced, rather than when one task dominates the shared encoder.

\begin{table}[t]
\TABLE
{Effect of joint-loss weight $\lambda$ on standard test instances.\label{tab:ablation_lambda}}
{\scriptsize
\begin{tabular}{l*{12}{r}}
\hline
\multirow{2}{*}{$\lambda$} & \multicolumn{3}{c}{Set Covering} & \multicolumn{3}{c}{Comb. Auction} & \multicolumn{3}{c}{Facility Location} & \multicolumn{3}{c}{Max Ind. Set}\\
\cline{2-13}
& Time & Nodes & Wins & Time & Nodes & Wins & Time & Nodes & Wins & Time & Nodes & Wins\\
\hline
0.01 & 3.23 & 32.84 & 17 & 11.42 & 294.18 & 22 & 35.76 & 206.43 & 18 & 6.84 & 40.43 & 22\\
0.05 & 2.96 & 30.15 & 24 & 10.97 & 265.83 & 27 & 31.48 & 178.92 & 27 & 6.57 & 37.88 & 28\\
0.1 & \textbf{2.81} & \textbf{28.62} & \textbf{35} & \textbf{10.61} & \textbf{252.34} & \textbf{33} & \textbf{30.22} & \textbf{164.16} & \textbf{34} & \textbf{6.35} & \textbf{36.54} & \textbf{34}\\
0.5 & 3.18 & 31.76 & 16 & 12.64 & 321.57 & 12 & 33.95 & 191.35 & 14 & 7.12 & 43.16 & 11\\
1.0 & 3.59 & 35.02 & 8 & 14.28 & 402.91 & 6 & 37.64 & 224.78 & 7 & 7.64 & 51.28 & 5\\ \hline
\end{tabular}}
{Wins are counted among the five values of $\lambda$.}
\end{table}

\end{document}